\documentclass[10pt,twocolumn,letterpaper]{article}

\usepackage[pagenumbers]{cvpr} 

\usepackage{graphicx}
\usepackage{amsmath}
\usepackage{amssymb}
\usepackage{booktabs}

\makeatletter
\@namedef{ver@everyshi.sty}{}
\makeatother
\usepackage{tikz}
\newcommand*\emptycirc[1][1ex]{\tikz\draw[thick] (0,0) circle (#1);}

\newcommand*\fullcirc[1][1ex]{\tikz\fill (0,0) circle (#1);}
\usepackage{booktabs}
\usepackage{multirow}
\newcommand{\ec}{\emptycirc[0.8ex]}

\newcommand{\fc}{\fullcirc[0.9ex]}

\usepackage{amsmath}
\usepackage{amsthm}
\usepackage{amsfonts}

\usepackage{bm}
\usepackage{xcolor}

\usepackage{url}
\usepackage{algorithm}
\usepackage[noend]{algorithmic}
\usepackage{graphicx}
\renewcommand{\algorithmiccomment}[1]{\bgroup\hfill $\triangleright$ ~#1\egroup}
\usepackage{hyperref}

\definecolor{newcolor}{rgb}{0.8,1,1}

\newcommand{\franknips}[1]{\textcolor{black}{#1}}

\newcommand{\vc}[1]{\textit{\textbf{#1}}}

\usepackage[capitalize]{cleveref}
\crefname{section}{Sec.}{Secs.}
\Crefname{section}{Section}{Sections}
\Crefname{table}{Table}{Tables}
\crefname{table}{Tab.}{Tabs.}

\def\cvprPaperID{*****} 
\def\confName{CVPR}
\def\confYear{2022}

\begin{document}

\title{$CV'3315$ Is All You Need -- Semantic Segmentation Competition}

\author{Akide Liu*\\
School of Computer Science\\
University of Adelaide\\
        {\tt\small a1743748@adelaide.edu.au }
\and
Zihan Wang*\\
School of Computer Science\\
University of Adelaide\\
{\tt\small a1810155@adelaide.edu.au}
}
\maketitle

\begin{abstract}
This competition \footnote{The University of Adelaide Course COMP SCI 3315 Computer Vision Semantic Segmentation Competition Lead by \href{https://irfanicmll.github.io/}{Dr. Yifan Liu}} focus on Urban-Sense Segmentation based on the vehicle camera view. Class highly unbalanced Urban-Sense images dataset challenge the existing solutions and further studies. Deep Conventional neural network-based semantic segmentation methods such as encoder-decoder architecture and multi-scale and pyramid-based approaches become flexible solutions applicable to real-world applications. In this competition, we mainly review the literature and conduct experiments on transformer-driven methods especially SegFormer \cite{xie2021segformer}, to achieve an optimal trade-off between performance and efficiency. For example, SegFormer-B0 achieved 74.6\% mIoU with the smallest FLOPS, 15.6G, and the largest model, SegFormer-B5 archived 80.2\% mIoU. According to multiple factors, including individual case failure analysis, individual class performance, training pressure and efficiency estimation, the final candidate model for the competition is SegFormer-B2 with 50.6 GFLOPS and 78.5\% mIoU evaluated on the testing set \footnote{Checkout our code implementation at \url{https://vmv.re/cv3315}}.
\end{abstract}

\section{Background}
\label{sec:intro}
Deep learning has been very successful when working with images as data and is currently at a stage where it works better than humans on multiple use-cases. The most critical problems humans have been interested in solving with computer vision are image classification, object detection and segmentation in the increasing order of their difficulty. In the plain old image classification task, people are just interested in getting the labels of all the objects present in an image. In object detection, researchers come further a step and try to know all objects in an image and the location at which the objects are present with the help of bounding boxes. Image segmentation takes it to a new level by trying to find out accurately the exact boundary of the objects in the image \cite{surr,minaee2021image}.

In this study, we, as beginners in the field of Computer Vision, aim to develop a basic understanding of semantic segmentation by reviewing, evaluating, and tuning existing methods, thereby providing a terrific solution that satisfies both efficiency and accuracy criteria to the given road segmentation task.

\begin{figure}[t]
  \centering
    \scalebox{0.55}{\includegraphics{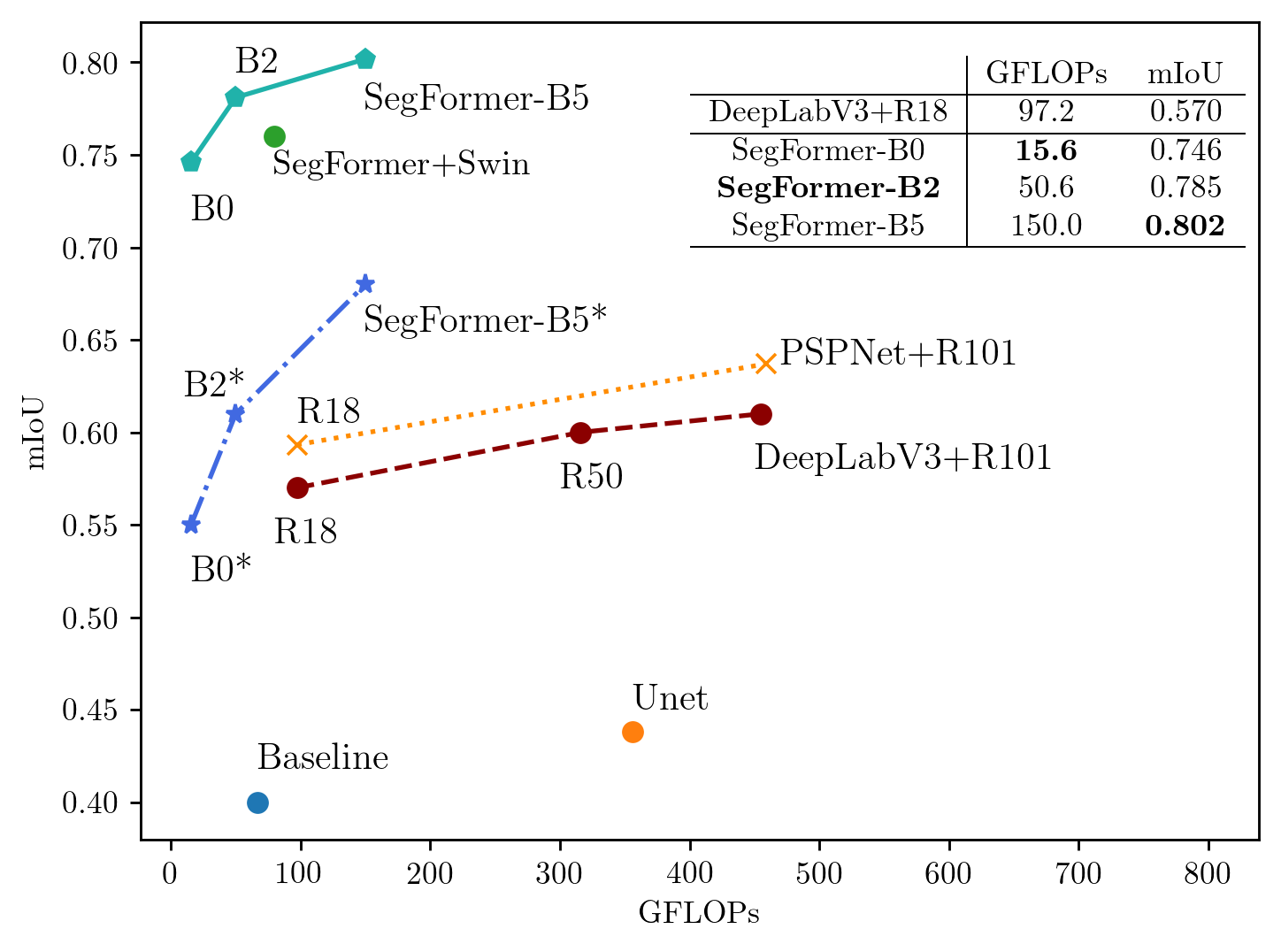}} 
    \caption{\textbf{Performance vs. model efficiency on the target dataset}. The symbol * indicates that the model did not pretrained on Cityscape dataset. ResNet and DeepLabV3Plus are abbreviated as R and DeepLabV3, respectively.}
    \label{fig:results}
\end{figure}

\subsection{Report Structure}
The remaining parts of this report are organized as follows. Section \ref{sec:method} reviews the baseline model and introduces methods and tricks to be applied. Experiments are conducted in Section \ref{sec:experiment}, and Section \ref{sec:conclusion} concludes the report.

\section{Method}
\label{sec:method}
\subsection{Baseline Model Analysis}
\begin{figure*}[t]
  \centering
    \scalebox{0.50}{\includegraphics{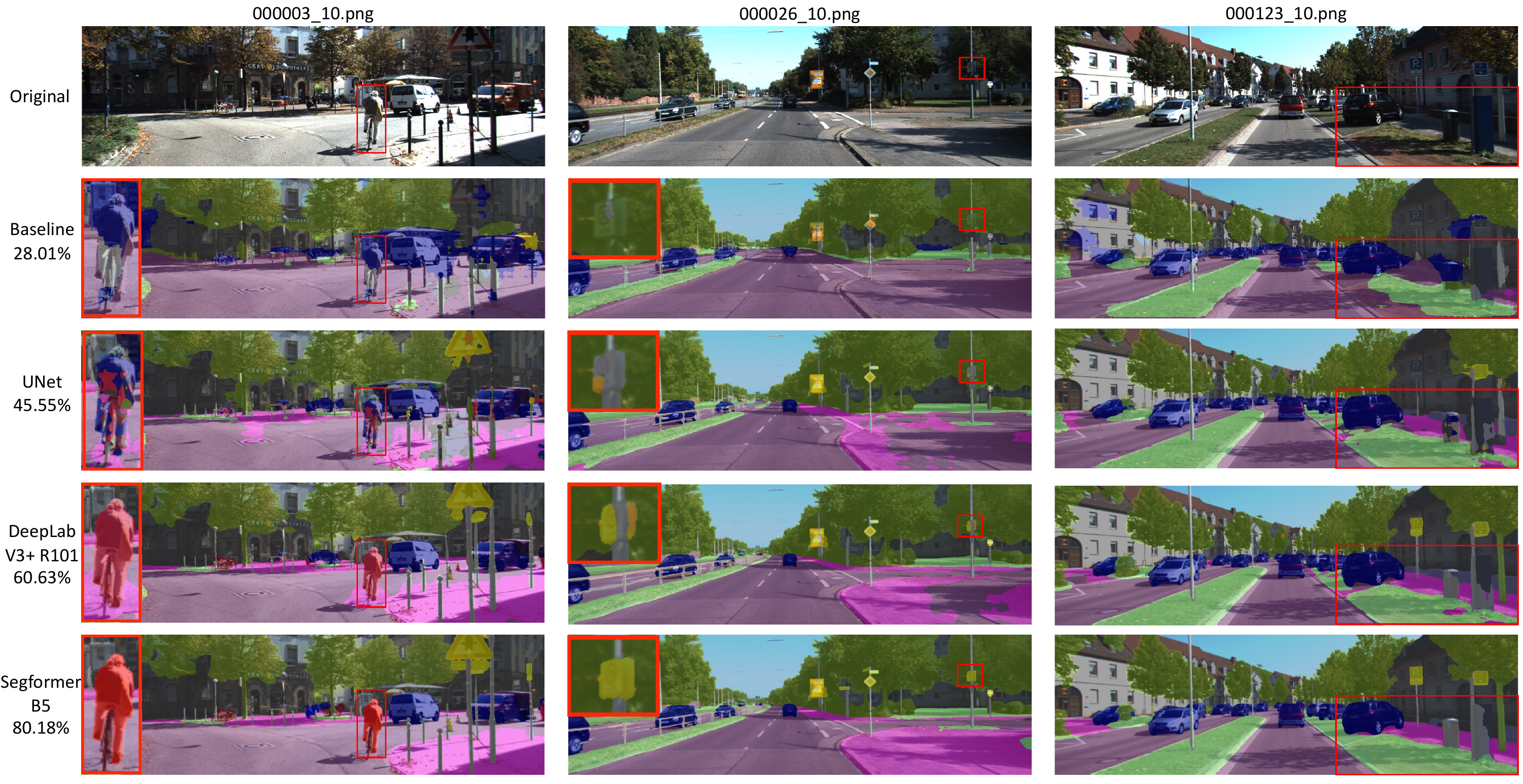}}
   \caption{\textbf{The visualization of segmentation results of test image}s. Compared to other models, SegFormer predicts masks with substantially finer details near object boundaries.}
   \label{fig:visualise}
\end{figure*}

The Baseline model comprises an Encoder-Decoder architecture. Basically, it extracts the feature maps from the image input and transforms them into the latent representation. The decoder network then retrieves those latent-space representations to perform predictions. Here, the latent-space representation refers to a high channel feature representation that combines useful underlying semantic information.

The encoder incorporates four downsampling layers to determine the intermediate features map. During the downsampling process, the activation function adopts RELU to improve the model's non-linearity. MaxPooling plays a significant role during the downsampling operations for spatial invariance; the pooling layer selects the maximum value of the current view. The convolutional layers take corresponding input channels [64, 128, 256, 512, 2048] with kernel size 3x3 and stride 2. The relatively small kernel decreases the number of parameters and also enhances the non-linearity; here, the stride is the moving step for the nearby convolution set to 2 to increase the receptive field.

The decoder of the baseline model plays an essential role in up-sampling the features map to the score map. The decoder network has four stages of up-sampling layers with a corresponding restore rate [1/16,1/8,1/4,1/2]. For each up-sampling layer using deconvolution as know as max-unpooling. Max-unpooling eliminates the need for learning to up-sample based on the location of the maximum value, the max-pooled values are placed, and a reminder of the matrix is loaded with zeros. The final layer directly connects to score maps using bilinear interpolation.

Also, we found that the baseline network is very similar to FCN, which is the fundamental work of semantic segmentation.

\textbf{FCN} \cite{fcn}. Long et al. first proposed using FCNs trained end-to-end for semantic segmentation. FCN utilizes a skip architecture that integrates semantic information from a deep, coarse layer with appearance information from a shallow, fine layer to produce accurate and detailed segmentations. FCNs have only locally connected layers, such as convolutions, pooling and upsampling, avoiding any densely connected layer. It also uses skip connections from its pooling layers to fully recover fine-grained spatial information lost during downsampling \cite{surr}.

\subsection{Networks Architecture Exploration}
\textbf{\underline{CNN.}} CNNs have been found quite efficacious for many computer vision tasks in recent years \cite{zhang2020resnest}. They act as trainable image filters that can convolve over images sequentially to measure responses or activations of the input image, creating feature maps. These feature maps are then piled together, passed through non-linear functions, and further convolved with more filters \cite{fcn}. This convolution process effectively extracts visual features or patterns in images that can be useful for tasks such as classification, segmentation, and super-resolution \cite{ gatescnn}. In this study, we first explore three CNN-based architectures for semantic segmentation: ResNet, U-Net and DeeplabV3, which we now review briefly.

\textbf{U-Net \cite{ronneberger2015u}}. U-Net is an encoder-decoder architecture that uses CNNs. Encoder-decoder networks, as the name suggests have two parts - an encoder and a decoder. The encoder is responsible for projecting the input feature vectors into a low-dimensional space in which similar features lie close together. The decoder network takes features from this low dimensional space as input and endeavours to reproduce the original input features. Thus, the output of the encoder or, conversely, the input of the decoder is called the bottleneck region where a low dimensional representation is present. Encoder-decoder networks have been found to be adequate for diverse tasks such as language translation and image segmentation.

\textbf{ResNet} \cite{zhang2020resnest}. The Deep Residual Nets (ResNet) model is initially proposed by He et al. to build a deep structure neural network for image classification. ResNet is a very popular DCNN model that won the annual ImageNet Large Scale Visual Recognition Challenge (ILSVRC) in 2015 for the classification task. The successfull achievement of ResNets model is then followed by other deep learning models as a framework to ease training of the respective model. We use three ImageNet pre-trained ResNet models as feature extractors of DeepLabV3 models.

\textbf{DeepLabV3} \cite{deeplab, deeplabv2}. DeepLabV3 utilizes atrous convolutions along with spatial pyramid pooling which enlarges the field of view of filters to incorporate a larger context and controls the resolution of features extracted. Employing atrous convolutions in either cascade or in parallel captures multi-scale context due to the use of multiple atrous rates. DeepLabV3 uses a backbone network such as a ResNet as its main feature extractor except that the last block is modified to use atrous convolutions with different dilation rates. This study explored the performance of DeepLabV3 as a semantic segmentation model using three ImageNet pre-trained ResNet models, namely: ResNet18, ResNet50, and ResNet101 \cite{he2016deep}.

Among all the CNN models mentioned above, the problems of spatial information extraction, edge reconstruction, and lightweight designs are solved to some extent; however, the global relations establishment problem is far from resolved. Therefore, we introduce transformers for segmentation.

\textbf{\underline{Transformer-based models.}}Transformer architecture is good at establishing global relations since the attention-mechanism-based designs compose the basic transformer unit but are less robust at extracting local information \cite{han2021transformer}. Therefore, the transformer architecture can perfectly resolve the four problems by referring to CNN designs. A shift from CNNs to transformers recently began with the Vision transformer (ViT). Basically, it treats each image as a sequence of tokens and then feeds them to multiple Transformer layers to make the classification. It is the first work to prove that a pure Transformer can achieve state-of-the-art performance in image classification \cite{pvt,vit}. 

Based on the pioneering work, many subsequent works made advancements. For instance, SETR \cite{setr} and TransUNet adopted skip connections to connect shallow and deep layers, surpassing CNN-based models in segmentation tasks. However, those Transformer-based methods have high computation complexity, damaging the application of the transformer in segmentation tasks \cite{surr}. 

Therefore, a lightweight transformer design is urgently needed. To this matter, SegFormer proposed a simple and efficient design without positional encoding and uses a simple decoder (segmentation head) to obtain the final segmentation results, thus, achieving a competitive performance to the models with the same complexity \cite{han2021transformer}. 

\textbf{SegFormer.} SegFormer is a simple, efficient yet powerful semantic segmentation framework which unifies Transformers with lightweight multilayer perceptron (MLP) decoders. SegFormer has two appealing features: 

\begin{enumerate}
  \item SegFormer comprises a novel hierarchically structured Transformer encoder which outputs multiscale features. It does not need positional encoding, thereby avoiding the interpolation of positional codes which leads to decreased performance when the testing resolution differs from training.
  \item It avoids complex decoders. The proposed MLP decoder aggregates information from different layers, and thus combining both local attention and global attention to render powerful representations.
\end{enumerate}

To better understand SegFormer, we analysed its network architecture:\\

Encoder -- Layered Transformer produces high resolution low-level features and low-resolution detail features.\\

Decoder -- A lightweight full MLP decoder fuses multi-level features to obtain semantic segmentation results.\\

\begin{figure}[t]
  \centering
    \scalebox{0.33}{\includegraphics{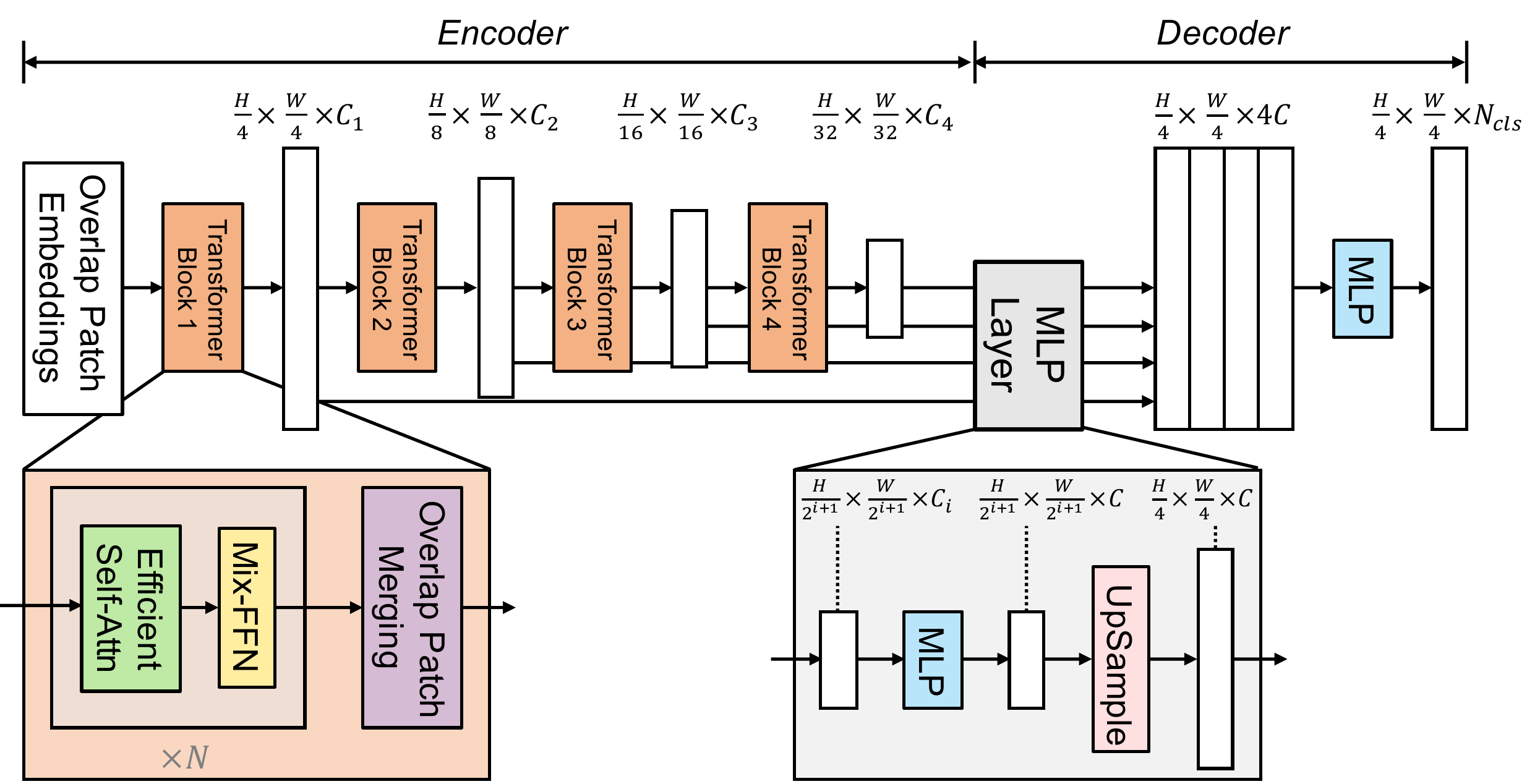}}  
   \caption{SegFormer \cite{xie2021segformer} framework.}
   \label{fig:segformer}
\end{figure}

\textbf{Hierarchical Transformer Encoder \cite{xie2021segformer}}
The author designed a series of MiT (Mix Transformer encoder) encoder models (MiT-B0 to B5), which have the same structure but different model sizes. The MiT design is inspired by ViT, but some optimizations have been made for semantic segmentation tasks.

\begin{itemize}
  \item \textbf{Hierarchical Feature Representation}. ViT can only generate single-resolution feature maps, similar to CNN's multi-scale feature maps, MiT generates feature maps of different scales, and the resolution of the generated feature maps is 1/4 1/8 1/16 1/32 of the original image.

  \item \textbf{Overlapped Patch Merging}. Patch Embedding of ViT is non-overlapping (non-overlapping), but non-overlapping will lead to discontinuous patch edges for semantic segmentation tasks. MiT uses overlapped patch embedding to ensure that the patch edges are continuous.

  \item \textbf{Efficient Self-Attention}. The main computational bottleneck of Transformer is the Attention layer. Let the dimension of Q, K, V be [N, C] (N=H*W), and the attention calculation can be formulated as follows:
\end{itemize}
\begin{equation}
\operatorname{Attention}(Q, K, V)=\operatorname{Softmax}\left(\frac{Q K^{\top}}{\sqrt{\operatorname{d}_{\text {head }}}}\right) V
\end{equation}

The complexity of the attention calculation is O($n^2$), When dealing with large-resolution images, SegFormer introduces a decay ratio R, and uses FC layer to reduce the amount of Attention calculation. The dimension of K is [N, C], what it does is to reshape it to [N/R, C*R] first, and then change the dimension to [N/R, C] through the fully connected layer, thereby reduce the computational complexity to O($N^2$/ R). R is set to [64, 16, 4, 1] from stage1 to stage4, respectively \cite{xie2021segformer}. 

\begin{equation}
    \begin{aligned}
&\hat{K}=\text { Reshape }\left(\frac{N}{R}, C \cdot R\right)(K) \\
&K=\operatorname{Linear}(C \cdot R, C)(\hat{K})
\end{aligned}
\end{equation}

\textbf{Mix-FFN (Feed-forward Network) \cite{xie2021segformer} }: The author believes that, instead of acquiring positional encoding,  information between adjacent pixels is more necessary for semantic segmentation tasks. A convolutional layer is applied in the MLP forward propagation (depthwise separable convolution is used to reduce the amount of parameters). The process can be formulated as follows:
\begin{equation}
    \mathbf{x}_{\text {out}}=\operatorname{MLP}\left(\operatorname{GELU}\left(\operatorname{Conv}_{3 \times 3}\left(\operatorname{MLP}\left(\mathbf{x}_{in \text {}}\right)\right)\right) ) \cdot \mathbf{x}_{\text {in}}\right.
\end{equation}
Experiments show that, 3x3 convolution can provide sufficient position information for the transformer.

\textbf{Lightweight All-MLP Decoder}. For feature maps $F_{i}$ of different resolutions, upsample them to 1/4, and then concat multiple feature maps and send them to the subsequent network to obtain segmentation results.

\begin{equation}
    \begin{aligned}&\hat{F}_{i}=\operatorname{Linear}\left(C_{i}, C\right)\left(F_{i}\right), \forall i \\&\hat{F}_{i}=\operatorname{Upsample}\left(\frac{W}{4} \times \frac{W}{4}\right)\left(\hat{F}_{i}\right), \forall i \\&F=\operatorname{Linear}(4 C, C)\left(\operatorname{Concat}\left(\hat{F}_{i}\right)\right), \forall i \\&M=\operatorname{Linear}\left(C, N_{c l s}\right)(F),\end{aligned}
\end{equation}
Note that SegFormer can be seen as a solid baseline for semantic segmentation since it achieves not only new state-of-the-art results on many datasets, but also shows strong zero-shot robustness. However, in this study, we only focus on its efficiency and performance, i.e. miou. Experiments have been conducted on 3 SegFormer models, namely \textbf{SegFormer-B0}, \textbf{SegFormer-B2} and \textbf{SegFormer-B5}.

\section{Experiments}
\label{sec:experiment}

\subsection{Dataset Description}

\textbf{Cityscapes.} The Cityscapes dataset \cite{cordts2016cityscapes} contains images from the driver’s perspective acquired in cities in Germany and neighbouring countries. The dataset provides 2MPx images split into Train, Val and Test subsets, where the semantic labels for the test subset are not publicly available. There are 19 label classes used for evaluation which we train upon. Train and Val subsets consist of 2,975 and 500 finely annotated images, respectively. We use this dataset for pre-train purposes.

\textbf{KITTI.} The KITTI semantic segmentation dataset \cite{Fritsch2013ITSC} was collected in Karlsruhe, Germany, while driving through the city and surrounding area. It provides 200 images for training and 200 images for testing at 1242x370px. The dataset uses the Cityscapes labelling policy.

\textbf{Target Dataset.} Contains 150 training images and 50 testing images. The given dataset is a subset of the KITTI semantic segmentation training set.

\begin{table*}\centering
\begin{tabular}{@{}l|cccc@{}}
\toprule
\textbf{Model Architecture}                            & \textbf{Pretrained Encoder} & \textbf{Pretrained Decoder} & \textbf{MIOU}   & \textbf{FLOPs}                    \\ \midrule
Baseline 180 Epochs Baseline                  & \ec                 & \ec                 & \textbf{0.2760} & \multirow{2}{*}{67.003G} \\
Baseline 1000 Epochs                          & \ec                 & \ec                 & 0.4271 &                          \\ \midrule
Unet                                          & \ec                 & \ec                 & 0.4555 & 356G                     \\ \midrule
Deeplabv3Plus \& ResNet18                     & \fc                & \ec                 & 0.5682 & 97.193G                  \\
Deeplabv3Plus \& ResNet50                     & \fc                & \ec                 & 0.6041 & 316G                     \\
Deeplabv3Plus \& ResNet101                    & \fc                & \ec                 & 0.6063 & 455G                     \\ \midrule
\multirow{2}{*}{Swin v1 Tiny 22k + SegFormer} & \fc                & \ec                 & 0.5945 & \multirow{2}{*}{79.472G} \\
                                              & \fc                & \fc                & 0.7628 &                          \\ \midrule
SegFormer B0*                                  & \fc                & \ec                 & 0.5533 & 15.579G                  \\
SegFormer B2*                                  & \fc                & \ec                 & 0.6141 & 50.633G                  \\
SegFormer B5*                                  & \fc                & \ec                 & 0.6842 & 150G                     \\ \midrule
SegFormer B0                                 & \fc                & \fc                & 0.7460 & \textbf{15.579}G                  \\
\textbf{SegFormer B2}                                 & \fc                & \fc                & 0.7845 & 50.633G                  \\
SegFormer B5                                 & \fc                & \fc                & \textbf{0.8018} & 150G                     \\ \midrule
PSPNet \& ResNet18                            & \fc                & \ec                 & 0.5933 & 97.213G                  \\
PSPNet \& ResNet101                          & \fc                & \ec                 & 0.6370 & 459G                     \\
CWD PSPNet \& R18(S) R101(T)                  & \fc                & \ec                 & 0.5449 & 99 G                     \\ \bottomrule
\end{tabular}
\caption{Overall Result. The symbol * indicates that the model did not pretrained on Cityscape dataset. \fc \space indicates that the model is pretrained and \ec \space means the opposite. Encoders are pretrained on ImageNet \cite{krizhevsky2012imagenet} and decoders are pretrained on Cityscape. CWD, T, S represent channel-wise distillation, Teacher and Student respectively. The last row is a future task and will be discuss in Section \ref{sec:conclusion}.}
\label{tab:bigtable}
\end{table*}

\subsection{Experiment Workflow}

We implement our experiments with PyTorch and MMSegmentation \cite{pytorch,mmseg2020} open-source toolbox. Most experiments are conducted on a server with 4 RTX A5000; we also configure and train models locally on one RTX 3080 Ti and one MPS device.

As a starting point, we directly train and run the baseline model with default settings for 180 epochs, and by evaluating, we obtain a mIoU of 0.2760, which is an abysmal result. To improve the model performance, we decided to follow the tricks in the notebook and see if a better result could be obtained. All results are displayed in \ref{tab:bigtable}.

Since we are dealing with a small dataset, data augmentation can be applied. We conducted multiple manipulations including RandomCrop, RandomFlip, PhotoMetricDistortion, Resize, Normalize and Pad to training samples \cite{shorten2019survey}; by training 1000 epochs, the mIoU increased from 0.3831 to 0.4271, demonstrating a significant performance improvement. Therefore we retain the argumentation strategy in all subsequent experiments. Following the hints, we slightly modify the baseline model architecture to a UNet, achieving a mIoU of 0.4613 for 1000 epochs. Since all hypermeters used so far are default, the next step is to tune the network. As it is known that the tuning process is more empirical than theoretical in deep learning, we did a board search and decided to conduct our experiments on an open-source toolbox — MMSegmentation.

MMSegmentation \cite{mmseg2020}. MMSegmentation is an open-source semantic segmentation toolbox based on PyTorch. It provides a unified benchmark toolbox for various semantic segmentation methods and directly supports popular and contemporary semantic segmentation frameworks, e.g. PSPNet, DeepLabV3Plus, SegFormer, etc. By decomposing the semantic segmentation framework into different components, the toolbox allows us to easily construct a customized semantic segmentation framework by combining different modules (e.g. Splice different encoder/decoder). Besides, it also supports many training tricks, and the training speed is faster than or comparable to other codebases since it supports DDP/DP. The toolbox enables us to tune/switch our model efficiently, quickly and intuitively and allows us to explore more experiments in a limited time. To ensure the dependability of the toolbox mmseg, we retested previous experiments, and precisely the same results are obtained by setting the Cuda seed unchanged (mIoU = 0.2760). Since the toolbox does not support our dataset, we wrote a data loader compatible with our dataset and the open community dataset KITTI. We are continuously updating the code with the support of MMSegmentation, aiming to contribute to the open-source community. Note that our source code can be reproduced in general practice, i.e. outside the MMSegmentation codebase.

To get a rough idea of how far we might be able to reach, we further explore the performance of the strong baseline DeepLabV3Plus as a semantic segmentation model using three ImageNet pre-trained Resnet models, namely: ResNet18, ResNet50, and ResNet101. Even with default hyperparameters, all three models achieve mIoU around 0.60, which is better but also shows that there is still room for improvement.

By looking at multiple SOTA ranking lists, we found that most top ones are transformer-based models. Therefore, we started to look into transformers. Taking into account the balance of efficiency and performance, we measure multiple models and eventually choose SegFormer — a simple, efficient and powerful semantic segmentation framework that unifies Transformers with lightweight multilayer perceptron (MLP) decoders. During the process of the development network, we have tried to switch SegFromer's original backbone to Swin Tiny Transformer \cite{liu2021swin} as it pre-trained on ImageNet 22K. After changing the architecture, the pre-trained model on cityscapes was obtained, which takes 8 hours via 4 Nvidia A5000 GPUs. The performance for Swin Tiny + SegFromer products is relatively high; however, the FLOPs of this modified architecture increased exponentially since this modified architecture will not consider for further optimization.

\subsection{Experiment with SegFormer (Useful tricks)}

The SegFromer contains six sizes of encoder and decoder to cover both efficiency and performance. We experimentally optimized hyperparameters and applied tricks on the most miniature model, Segformer-B0, which is suitable for faster development and reduces computational cost on our local machine. After optimal hyper-parameter or efficient tricks are employed in the model, we duplicate the exact strategies to the larger models (B2 and B5) and push them to cloud servers to perform distributed training and evaluation. This workflow allows large batch sizes, large crop sizes, sync batch normalization and multiple scales training/testing to improve model performance; furthermore, distribution data parallelism, multiple GPUs cross gradient descent to reduce training duration.

\subsubsection{Transfer learning}

Transfer learning is a machine learning technique where a model trained on one task as an optimization allows rapid progress or improved performance when modelling the second task \cite{weiss2016survey}. After a manual examination of the dataset properties and attributes, we found that this dataset belongs to Urban-scene segmentation tasks, and the image size is 1242x375 with 19 classes appropriately labelled. According to these clues and some online research, more detailed descriptions and information have been allocated. This dataset is a subset of the KITTI semantic segmentation benchmark, containing 200 semantically annotated train images, and the data format conforms with \textbf{Cityscapes} Dataset. Note that we did not use the Original KITTI dataset as data enhancement for this competition because our testing set is also the subset of the KITTI training set to avoid biased performance. Some research shows that models trained on a large dataset such as Cityscapes have well generalization ability for unseen data. Therefore, we decided to use Cityscapes pre-trained weights to consolidate our task \cite{kim2017end} . The reuse model is a methodology for inductive transfer learning; the Cityscapes pre-trained model is used as the starting point instead of random initialization or batches normal initialization. \textbf{After employing transfer learning, the model performance increased from 61.41\% to 68.31\% mIoU.}

\subsubsection{Learning Rate Scheduler}

Some research shows that the learning rate is one of the most important hyperparameters for deep learning because it directly affects the gradient descents and controls the speed of network convergence to the point of global minima by navigation through a non-convex loss surface \cite{ding2019adaptive}. How find an optimal learning rate becomes a huge challenge for our experiments. After several experiments, we used a polynomial learning rate policy with a warm restart. The following equation defines the polynomial learning rate for the task of semantic image segmentation. Here, $lr_0$ is the initial or base learning rate, $i$ denotes the number of iterations, and $T_i$ is the total number of iterations. In case of polynomial learning rate policy, $T_i$ is equal to the total number of epochs times number of iterations in an epoch. Power term controls the shape of learning rate decay. Our optimal parameters are $lr0 = 1e-5, power = 1.0, T_i = 20000$. Furthermore, use AdamW optimizer  , which leverages the L2 normalization and weight decay with Adam optimizer. \textbf{After employing a learning rate scheduler, the model performance increased from $68.31\%$ to $72.34\%$ mIoU.}

\begin{equation}
    \begin{aligned}
    l r=l r_{0} *\left(1-\frac{i}{T_{i}}\right)^{\text {power }}
    \end{aligned}
\end{equation}

\def\R{{\rm I\hspace{-0.50ex}R}}

\begin{algorithm}[tb!]
\caption{Adam with L$_2$ regularization and Adam with \franknips{decoupled} weight decay (AdamW)}
\footnotesize
\label{algo_adam}
\begin{algorithmic}[1]
\STATE{\textbf{given} $\alpha = 0.001, \beta_1 = 0.9, \beta_2 =0.999, \epsilon = 10^{-8}, \lambda\in \R$} \label{adam-Given}
\STATE{\textbf{initialize} time step $t \leftarrow 0$, parameter vector $\bm{\theta}_{t=0} \in \R^n$,  first moment vector $\vc{m}_{t=0} \leftarrow \vc{0}$, second moment vector  $\vc{v}_{t=0} \leftarrow \vc{0}$, schedule multiplier $\eta_{t=0} \in \R$}
\REPEAT
	\STATE{$t \leftarrow t + 1$}
	\STATE{$\nabla f_t(\bm{\theta}_{t-1}) \leftarrow  \text{SelectBatch}(\bm{\theta}_{t-1})$}  \COMMENT{select batch and return the corresponding gradient}
	\STATE{$\vc{g}_t \leftarrow \nabla f_t(\bm{\theta}_{t-1})$  $+ \lambda\bm{\theta}_{t-1}$}
	
	\STATE{$\vc{m}_t \leftarrow \beta_1 \vc{m}_{t-1} + (1 - \beta_1) \vc{g}_t $} \label{adam-mom1} \COMMENT{here and below all operations are element-wise}
	\STATE{$\vc{v}_t \leftarrow \beta_2 \vc{v}_{t-1} + (1 - \beta_2) \vc{g}^2_t $} \label{adam-mom2}
	\STATE{$\hat{\vc{m}}_t \leftarrow \vc{m}_t/(1 - \beta_1^t) $} \COMMENT{$\beta_1$ is taken to the power of $t$} \label{adam-corr1}
	\STATE{$\hat{\vc{{v}}}_t \leftarrow \vc{v}_t/(1 - \beta_2^t) $} \COMMENT{$\beta_2$ is taken to the power of $t$} \label{adam-corr2}
	\STATE{$\eta_t \leftarrow \text{SetScheduleMultiplier}(t)$}	\COMMENT{can be fixed, decay, or also be used for warm restarts}
	\STATE{$\bm{\theta}_t \leftarrow \bm{\theta}_{t-1} - \eta_t \left( \alpha  \hat{\vc{m}}_t / (\sqrt{\hat{\vc{v}}_t} + \epsilon) + \lambda\bm{\theta}_{t-1} \right)$} \label{adam-xupdate}
\UNTIL{ \textit{stopping criterion is met} }
\RETURN{optimized parameters $\bm{\theta}_t$}
\end{algorithmic}
\end{algorithm}




\subsubsection{Class Balanced Loss \cite{cui2019class}}

\begin{figure}[t]
  \centering
    \scalebox{0.2}{\includegraphics{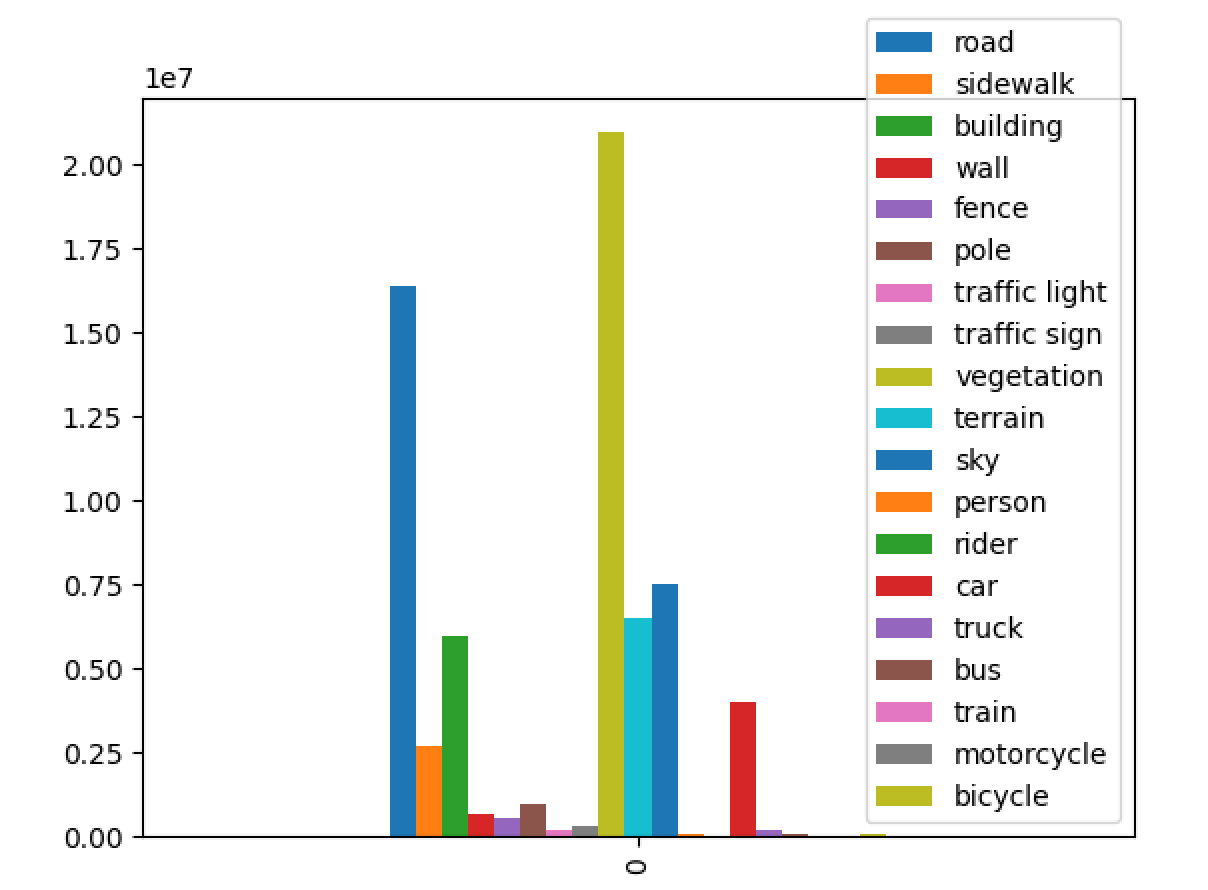}}  
   \caption{Class distribution diagram of the target dataset.}
   \label{fig:class_distribution}
\end{figure}

\begin{itemize}
  \item The urban-scene images have distinct nature related to perspective geometry and positional patterns. Since the urban-scene images are captured by the cameras mounted on the front side of a car, the urban-scene datasets consist only of road-driving pictures. This leads to the possibility of incorporating common structural priors depending on a spatial position, markedly in a vertical position. The class distribution diagram of the target dataset is shown in \ref{fig:class_distribution}.
  \item After manually examining the dataset, we write a simple pixels distribution analyzer for visualization and calculate the class balanced loss. The collected class balanced weights added to CrossEntropyLoss generate a class balanced loss.
  \item A re-weighting scheme uses an adequate number of samples for each class to re-balance the loss, yielding a class-balanced loss.
  \item The class-balanced loss is designed to address the problem of training from imbalanced data by introducing a weighting factor that is inversely proportional to the adequate number of samples.
  \item \textbf{After employing Class Balanced Loss, the model performance increased from 72.34\% to 74.67\% mIoU.}

\end{itemize}

\begin{equation}
  \mathrm{CE}_{\mathrm{softmax}}(\mathbf{z}, y)=-\log \left(\frac{\exp \left(z_{y}\right)}{\sum_{j=1}^{C} \exp \left(z_{j}\right)}\right)
\end{equation}

\subsubsection{Online Hard Example Mining (OHEM) \cite{ohem}}

\begin{itemize}
  \item The OHEM algorithm (online hard example miniing, CVPR‘16) is mainly for automatic selection of difficult samples in the training process. larger samples), and then apply these filtered samples to training in stochastic gradient descent.
  \item Only pixels with confidence score under thresh are used to train. And we keep at least $min_kept$ pixels during training. If thresh is not specified, pixels of top $min_kept$ loss will be selected. The optimal parameters for OHEM are $thresh = 0.5, min_{kept}=10k$
  \item \textbf{After employing OHEM the model performance increased from 74.64\% to 75.98\% mIoU.}
\end{itemize}

\subsubsection{Multiple Scale Testing \cite{ren2016object}}

\begin{figure}[t]
  \centering
    \scalebox{0.35}{\includegraphics{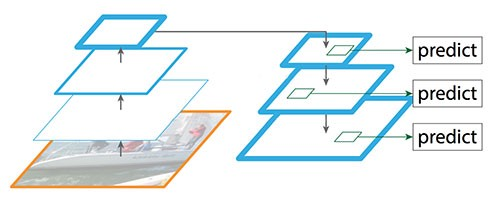}}  
   \caption{Multiple Scale Testing}
   \label{fig:multiple_test}
\end{figure}

The multiple scale testing generates the input images and features maps to different scales, determines the region proposal and combines the region proposal for mixed scale input to the network evaluation. The multiple scales are widely used in object detection to reduce the detection difficulty for some small targets. For example, the SSD makes detection from multiple feature maps; however, the bottom layers are not selected for object detection. They are in high resolution, but the semantic value is not high enough to justify their use as the speed slow-down is significant. So SSD only uses upper layers for detection and performs much worse for small objects. This method has been implemented in our codebase, creating multiple-scale images input for prediction and averaging the scores as the final prediction. \textbf{After employing multiple-scale testing/training, the model performance increased from 75.98\% to 78.12\% mIoU.}

\subsubsection{Auxiliary Loss \cite{pspnet}}

As the network becomes deeper, the training progress decreases because of vanishing gradient issues. To reduce the impact, auxiliary loss with suitable weights helps optimize the learning process, while the main branch loss takes the most responsibility. In our experiment, we added the FCN layer as an auxiliary head to interrupt the second intermediate layer to make predictions and calculate the auxiliary loss. Then merge auxiliary loss with main loss from decoder head. In our implementation, the auxiliary loss weight is 0.4 against the main loss weight of 1.0. \textbf{After employing Auxiliary loss, the model performance increased from 78.12\% to 78.45\% mIoU.}

\section{Conclusion and Future Prospects}
\label{sec:conclusion}
Overall, by applying the strong baseline SegFormer we achieve fantastic results. The most efficient-performance balanced model SegFormer-B2 (FLOPs = 79.472G), a mIoU of 0.7845 is obtained. With the strongest SegFormer-B5, mIoU = 0.8018 is reached with FLOPs = 150G. A great model and thoughtful tuning are essential to produce the best performance. Transfer learning is also very crucial in improving model performance, and we will especially focus on exploring this in the future. During this study, we develop a comprehensive understanding of the field of deep learning and semantic segmentation. In addition to the joy of harvesting experiment results, we also develop strong interests in computer vision. The fly in the ointment is that we are not able to compose more work due to the time limit — we tried to apply knowledge distillation (channel-wise distillation)\cite{shu2021channel} to our model, and relevant experiments have been successfully conducted on a PSPNet model \cite{pspnet}. We will keep working on this; ideally, we can successfully employ it on the top-performing model within two weeks.

{\small
\bibliographystyle{ieee_fullname}
\bibliography{egbib}
}

\end{document}